%% file: main.tex
\documentclass{article}
\usepackage{spconf}
\input{math_commands.tex}

\usepackage{hyperref}
\usepackage{easy-todo}
\usepackage{graphicx}
\usepackage{amsmath,amssymb,amsthm}
\usepackage{tikz}
\usepackage{forloop}
\usepackage{makecell}

\title{Dynamic Variational Autoencoders for Visual Process Modeling}
\name{Alexander Sagel\sthanks{\texttt{sagel@fortiss.org}. Alexander Sagel carried out a part of the research during his stay at Inria in Rennes, France.}, \quad Hao Shen\sthanks{\texttt{shen@fortiss.org}}}
\address{fortiss - The Research Institute of the Free State of Bavaria\\Munich, Germany}
\begin{document}

%
\maketitle
\begin{abstract}
	This work studies the problem of modeling visual processes by leveraging deep 
	generative architectures
	for learning linear, Gaussian representations
	from observed sequences.
	We propose a joint learning framework, combining a \emph{vector autoregressive model} and a \emph{Variational Autoencoder}.
	This results in an architecture that allows Variational
	Autoencoders to simultaneously learn a non-linear observation 
	as well as a linear state model from sequences of frames. 
	We validate our approach on synthesis of artificial sequences and dynamic textures. To this end, we use our architecture to learn a statistical model of each visual process, and generate a new sequence from each learned visual process model.
\end{abstract}
\section{Introduction}
Many visual real-world phenomena such as motion captures of bird migrations, crowd movements or ocean waves can be thought of as random processes. From a statistical point of view, these \emph{visual processes} can be thus described in terms of their probabilistic properties, if an appropriate model can be inferred from observed sequences of the process.
Such a model can be helpful for studying the dynamic properties of the process, estimating possible trajectories \cite{Johnson2016}, detecting anomalous behavior \cite{Mahadevan2010} or generating new sequences \cite{Vondrick2016}.

One of the most classic models for visual processes is the linear dynamic system (LDS) based approach proposed by Doretto et. al \cite{Doretto2003}. It is a holistic and generative sequential \emph{latent variable} description. Since it is essentially a combination of linear transformations and additive Gaussian noise, it is mathematically simple and easy to interpret. 
However, it bears the disadvantage of constraining each video frame to lie in an low-dimensional affine space. Since real-world observation manifolds are anything but linear, this restricts the applicability to very simple visual processes.

The contribution of our work is to generalize the LDS model by Doretto et al. to  non-linear visual processes, while still keeping its advantages of being an easy-to-analyze latent variable model. To this end, we use Variational Autoencoders to learn a non-linear state space model, in which the latent states still follow a linear, autoregressive dynamic. This is done by supplementing the Variational Autoencoder architecture with a linear layer that models the temporal transition.

%
%
%

\section{Background}
\label{sec:related}
The dynamic texture model in \cite{Doretto2003} is an LDS of the form
\vspace{-1mm}
\begin{equation}
\vh_{t+1}= \mA \vh_{t}+ \vv_t,\qquad
\vy_{t}=\bar{ \vy}+ \mC \vh_t,
\label{eq:dyntex}
\end{equation}
where $ \vh_t\in\R^n,\ n<d,$ is the low dimensional \emph{state space variable} at time $t$, $ \mA\in\R^{n\times n}$ the \emph{state transition matrix}, $ \vy_t\in\R^{d}$ the observation at time $t$, and $\mC\in\R^{d\times n}$ the full-rank \emph{observation matrix}. The vector $\bar{ \vy}\in\R^d$ represents a constant offset of the observation space and the input term $ \vv_t$ is modeled as i.i.d. zero-mean Gaussian noise that is independent of $ \vh_t$. Learning a model as described in \eqref{eq:dyntex} is done by inferring $\mA$ and $\mC$ from one or several videos. To do so, it is sensible to assume that the process is (approximately) second-order stationary and first-order Markov. We make the same two assumptions, although our method is easily generalizable to Markov processes of higher orders.

Real-world visual processes are generally non-linear and non-Gaussian. Nevertheless, the model \eqref{eq:dyntex} has considerable appeal in terms of simplicity. This is particularly true for the latent model of $\vh_t$ in the system equation which consists of a vector autoregressive (VAR) process. For instance, while it is theoretically possible to replace the latent VAR by a recurrent neural network (RNN) with noise input to obtain a much richer class of temporal progressions, such an adjustment would lead to a loss of tractability, due to the unpredictable long-term behavior of RNNs as opposed to linear transition models. Modeling the expected temporal progression as a matrix multiplication also greatly simplifies the inversion and interpolation of a temporal step by means of matrix algebra. Luckily, we can still generalize the system \eqref{eq:dyntex} while keeping the latent VAR model by employing  
\begin{equation}
\begin{split}
\vh_{t+1}= \mA \vh_{t}+ \vv_t,\qquad
\vy_{t}=C( \vh_t),
\label{eq:dyntex_nl}
\end{split}
\end{equation}
where $n<d$ and $C \colon \R^n\to\R^d$ is a function which maps to the observation manifold $\sM$, i.e. the manifold of the frames produced by the visual process, and is bijective on $C^{-1}(\sM)$.

We propose to keep the latent VAR model and combine it with a \emph{Variational Autoencoder} (VAE) \cite{Kingma2013} to generate video frames on a non-linear observation manifold from the VAR samples. Since it would go beyond the scope of this paper to review VAEs, we refer the reader to \cite{Doersch2016} and make do with the simplification that the VAE is an autoencoder-like deep architecture that can be trained such that the decoder maps standard Gaussian noise to samples of the probability distribution of the high-dimensional training data. This is done by optimizing the so-called \emph{variational lower bound}. 

While it is trivial to learn a deep generative model such as a VAE, and combine it with a separately learned VAR process, we propose to learn both aspects jointly, to fully exploit the model capacity. One can thus describe our method as system identification \cite{Vanovershee1994} with deep learning.

Combinations of latent VAR models with learned, non-linear observation functions have been explored in previous works. For instance, in \cite{Chan2007} the authors apply the kernel trick to do so. Approaches of combining VAR models with deep learning have been studied in \cite{Chalasani2013, Liu2018}, but neither of these employ VAEs. By contrast, the authors of \cite{Watter2015} combine Linear Dynamic Systems (LDSs) with VAEs. 
However, their model is \emph{locally} linear and the transition distribution is modeled in the variational bound, whereas we model it as a separate layer. This also is the main difference to the work \cite{Krishnan2015}, in which VAEs are used as Kalman Filters, and the work \cite{Karl2016} which uses VAEs for state space identification of dynamic systems. Similarly, the work \cite{Johnson2016} combines VAEs with linear dynamic models for forecasting images but proposes a training objective that is considerably more complex than ours.


From an application point of view, generative modeling of visual processes is closely related to \emph{video synthesis} which has been discussed, among others in \cite{Vondrick2016,Xue2016,Srivastava2015}. 
Two of the most recent works in the generative modeling and synthesis of visual processes are \cite{Xie2017,Tesfaldet2018}. Both models achieve very good synthesis results for dynamic textures. One advantage of our model over these techniques is the ability to synthesize videos frame-by-frame in an online manner once the model is trained, without significantly increasing memory  and time consumption as the sequence to be synthesized becomes longer. This is due to our model not requiring an optimization procedure for each synthesized sequence.

\section{Joint Learning via Dynamic VAEs}
\label{sec:model}
In the following, we describe our method to learn a model described by \eqref{eq:dyntex_nl} from video sequences.
We use upright characters, e.g. $\rvx, \rvy$ to denote random variables, as opposed to italic ones, e.g. $\vx, \vy$ that we use for all other quantities.

Without loss of generality \cite{Afsari2013,Sagel2017}, we assume that the marginal distribution of each state space variable $\rvh_t$ is standard Gaussian at any time $t$. Thus, the model in \eqref{eq:dyntex_nl} is entirely described by the matrix $\mA$ and the function $C$.
A sequence of length $N$ of a visual process $\mathcal Y$ is viewed as a realization of the random variable $\rvy^N=[\rvy_1,\dots,\rvy_N]$. The according sequence in the latent space is a realization of the random variable $\rvh^N=[\rvh_1,\dots,\rvh_N]$. Let us define the random variable 
\begin{equation}
\tilde{\rvy}^N=[\tilde{\rvy}_1,\dots,\tilde{\rvy}_N]:=[C(\rvh_1),\dots,C(\rvh_N)].
\end{equation}
In order to model $\mathcal{Y}$ by \eqref{eq:dyntex_nl}, we need to make sure that the probability distributions of $\rvy^N$ and $\tilde{\rvy}^N$ coincide for any $N \in \mathbb{N}$. Taking into account the stationarity and Markov assumptions from \secref{sec:related}, this is equivalent to demanding that the joint probabilities of two succeeding frames coincide.

The joint probability distribution of two succeeding states $\rvh_t$ and $\rvh_{t+1}$ is zero-mean Gaussian and, the following relation holds.
\begin{equation}
\begin{bmatrix}\rvh_t\\ \rvh_{t+1}\end{bmatrix} \sim \mathcal N\left(\begin{bmatrix}\rvh_t\\ \rvh_{t+1}\end{bmatrix};\ 0,\begin{bmatrix}
\mI & \mA^\top\\
\mA &\mI
\end{bmatrix}\right) \ \forall t.
\label{eq:h2}
\end{equation}

With this in mind, the problem boils down to finding a function $C$ and a matrix $\mA$ such that the random variable $\tilde{\rvy}^2=\begin{bmatrix}C(\rvh_1)&C(\rvh_2)\end{bmatrix}$
has the same distribution as $\rvy^2$.

Since a VAE decoder can be trained to map from a standard Gaussian to a data distribution, it makes sense to employ it as the observation function
$C$. Unfortunately, a classical VAE does not provide a framework to simultaneously learn the matrix $\mA$ from sequential data. However, we still can use a VAE to accomplish this task. Let us denote by $f_\theta$ a successfully trained VAE decoder, i.e. a function that transforms samples $\vx$ of standard Gaussian noise to samples from a high dimensional data distribution. The variable $\theta$ denotes the entirety of trainable parameters of $f_\theta$. Let us assume that $\theta=(\mA,\mB,\eta)$ contains the matrices $\mA,\mB\in\R^{n\times n}$ such that
\begin{equation}
\mA\mA^\top+\mB\mB^\top=\mI_n
\label{eq:st_p}
\end{equation} 
is satisfied, and the weights $\eta$ of the subnetwork $C_\eta$ of $f_\theta$ that implements the observation function $C$. Consider the following definition for $f_\theta$.
\begin{equation}
f_\theta:\R^{2n}\to\R^{2d}, \qquad
\begin{bmatrix}
\vx_1\\\vx_2
\end{bmatrix}\mapsto
\begin{bmatrix}
C_\eta\left(\vx_1 \right) \\
C_\eta\left(\mA\vx_1+\mB \vx_2 \right)
\end{bmatrix}.
\label{eq:ftheta}
\end{equation}
If $f_\theta$ is trained to map $\rvx\sim\mathcal{N}(\rvx;0,\mI_{2n})$ to $\tilde{\rvy}^2=f_\theta(\rvx)$ with 
$\tilde{\rvy}^2\sim p(\rvy^2)$,
then $\mA$ and $C=C_\eta$ indeed fulfill the property that the joint probabilities of the random variables $\tilde{\rvy}^2$ and ${\rvy}^2$, as defined above, coincide, if the random variables
$\rvh_1,\rvh_2$ follow the joint probability distribution in \eqref{eq:h2}.

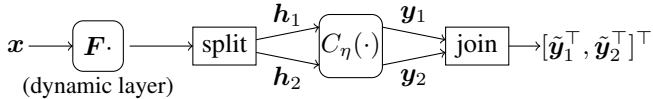
\begin{figure}
	\begin{center}
		\begin{tikzpicture}[scale=0.84]
		\draw[->] (-4.6,0)-- (-3.9,0);
		\draw[->] (-1,0.1)-- (0,0.3);
		\draw[->] (-1,-0.1)-- (0,-0.3);
		\draw[rounded corners] (-3.9, -0.4) rectangle (-3.1,0.4);
		\node (F) at (-3.5,0) {$\mF \cdot$};				\node (dl) at (-3.5,-0.65) {\small{(dynamic layer)}};
		
		\node (C) at (0.5,0) {$C_\eta(\cdot)$};
		\node (x) at (-4.8,0) {$\vx$};
		\node (h1) at (-0.5,0.5) {$\vh_1$};
		\node (h2) at (-0.5,-0.5) {$\vh_2$};
		\node (h1) at (1.5,0.5) {$\vy_1$};
		\node (h2) at (1.5,-0.5) {$\vy_2$};
		\node (split) at (-1.5,-0) {split};
		\draw[rounded corners] (0, -0.5) rectangle (1,0.5);
		\draw (-2, -0.3) rectangle (-1,0.3);
		\draw[->] (1,0.3)-- (2,0.1);
		\draw[->] (1,-0.3)-- (2,-0.1);
		\draw[->] (-3,0)-- (-2,0);
		\draw (2, -0.3) rectangle (3,0.3);
		\node (join) at (2.5,-0) {join};
		\draw[->] (3,0)-- (3.5,0);
		\node (x) at (4.4,0) {$[\tilde\vy_1^\top,\tilde\vy_2^\top]^ \top$};
		\end{tikzpicture}	
	\end{center}
	\caption{Decoder of a Dynamic VAE with a dynamic layer.}
	\label{fig:vae}
\end{figure}

To model $f_\theta$ as a neural network, we formalize the initial transformation of $\vx$ in \eqref{eq:ftheta} as a multiplication with 
\begin{equation}
\mF=\begin{bmatrix}\mI_n & \boldsymbol{0}\\ \mA & \mB \end{bmatrix}
\end{equation}
from the left. The function $f_\theta$ can be realized by the neural architecture depicted in \figref{fig:vae}. The first layer is linear and implements the multiplication with $\mF$. We refer to this layer as the \emph{dynamic layer}. The output of the dynamic layer is divided into into an upper half $\vh_1$ and a lower half $\vh_2$ and both halves are fed to the subnetwork that implements $C_\eta$. The weights of the dynamic layer contain the matrices $\mA,\mB$. Thus, they can be trained along $\eta$, by back-propagation of the stochastic gradient computed from pairs of succeeding video frames. When the architecture is employed as the decoder of a VAE, the parameters $\mA$ and $\eta$ are implicitly trained to satisfy the desired requirements. 
However, we need to make sure that the stationarity constraint in \eqref{eq:st_p} is not violated. We have observed that this can be effectually done by adding a simple regularizer to the loss function of the VAE. Let $L(\theta,\vartheta)$ denote the variational lower bound of the VAE, with $\theta, \vartheta$ denoting the trainable parameters of the decoder and encoder respectively. The loss function of our model is given as
\begin{equation}
\tilde{L}(\theta,\vartheta)=L(\theta,\vartheta)+\lambda\|\mA\mA^\top+\mB\mB^\top-\mI_n\|_F^2,
\end{equation}
where $\lambda>0$ is a regularizer constant. We refer to the resulting model consisting of a VAE with a dynamic layer and a stationarity regularizer, as the \emph{Dynamic VAE} (DVAE).
\section{Experiments}
One way to evaluate a generative video model is to generate synthetic sequences of visual processes from the trained model and to assess how much the their statistics correspond to the training data. Unfortunately, most quantitative quality measures that are common in video prediction such as Peak Signal to Noise Ratio or  Structural Similarity are unsuitable for evaluating video synthesis since they favor entirely deterministic models. As a more appropriate quality measure for generative models, the \emph{Fr\'echet Inception Distance} (FID) \cite{Heusel2017} has been widely accepted.

However, even the FID is designed to compare the probability distributions of  isolated still-image frames, neglecting the temporal behavior of the process. Additionally to the FID score, it is thus necessary to visually inspect the synthesis results. In this section we thus aim to do both. \subref{sub:expvae} provides qualitative results on artificial sequences in order to demonstrate the inference capabilities of our model. \subref{sub:dyntex} additionally provides quantitative comparisons with state-of-the-art methods for dynamic texture synthesis, in particular the recent \emph{Spatial-Temporal Generative Convet} (STGCONV) \cite{Xie2017} and the \emph{Dynamic Generator Model} \cite{Xie2018}, whenever the results were made available. To this end, FID scores are computed via \cite{FIDCODE}.

Synthesis is performed by sampling from 
\begin{equation}
\vh_{t+1}=\mA\vh_t+\mB\vv_t,\qquad \rvv_t \sim\  \mathrm{i.i.d.}\  \mathcal N(\rvv_t,0,\mI_n),
\end{equation}
and mapping the latent states $\vh_t$ to the observation space by means of $C_\eta$. The initial latent state $\vh_0$ is estimated by applying the encoder to a frame pair from the training sequence. We train $C_\eta$ and $\mA$ simultaneously by means of our Dynamic VAE framework introduced in \secref{sec:model}. As the encoder of the VAE, we use the discriminator of the DCGAN, but adapt the number of output channels to be $2n$, where $n$ is the latent dimension of the model. As the decoder, we employ the DCGAN generator. The latent dimension is set to $n=10$ for all  experiments. The number of convolutional layers and the size of the convolutional kernel of the decoder output layer (encoder input layer) varies to match the resolution of the input data. We set the regularizer constant to $\lambda = 100$. The exact configuration is listed in Table~\ref{tab:conf}, where $\sigma_\rvy^2$ denotes the conditional variance of the VAE output distribution. PyTorch code is available at \cite{DVAECODE}.

\begin{table}
	\begin{center}
		\begin{tabular}{c c c c c c}
			\thead{Experiment} & \thead{Resolution}  & \thead{Conv.\\layers} & \thead{Kernel\\size} & \thead{$\sigma_\rvy^2 $} \\
			\hline
			MNIST & $32\times 32$ &4&4&8.0 \\
			Running Cows & $64\times 64$ &5&4&10.0 \\
			Salt+Pepper mask & $64\times 64$ &5&4&4.5 \\
			Rectangular mask & $128\times 128$ &5&8&8.0 \\
			Dyn. Textures & $128\times 128$ &5&8&16.0
		\end{tabular}
	\end{center}
	\caption{Experimental Configuration}
	\label{tab:conf}
\end{table}

\subsection{Transformation Sequences}
\label{sub:expvae}
\newcounter{imagenumber}
\setlength{\tabcolsep}{0.3pt}

\begin{figure}
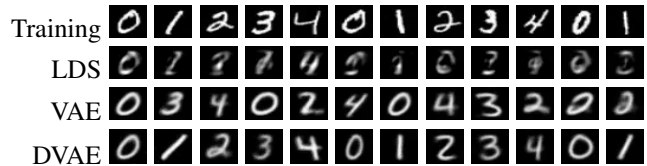

	\begin{center}
		\begin{tabular}{rccccccccccccccc}	
			Training 
			\forloop{imagenumber}{1}{\value{imagenumber} < 10}{
				& \includegraphics[width=14pt, height=12pt]{results/original/mnist/mnist_01234/image000\arabic{imagenumber}.png}
			}\forloop{imagenumber}{10}{\value{imagenumber} < 13}{
				& \includegraphics[width=14pt, height=12pt]{results/original/mnist/mnist_01234/image00\arabic{imagenumber}.png}
			} \\
			
			LDS 
			\forloop{imagenumber}{1}{\value{imagenumber} < 10}{
				& \includegraphics[width=14pt, height=12pt]{results/linear/mnist_linear/mnist_01234/image000\arabic{imagenumber}.png}
			}\forloop{imagenumber}{10}{\value{imagenumber} < 13}{
				& \includegraphics[width=14pt, height=12pt]{results/linear/mnist_linear/mnist_01234/image00\arabic{imagenumber}.png}
			} \\
			
			VAE
			\forloop{imagenumber}{1}{\value{imagenumber} < 10}{
				& \includegraphics[width=14pt, height=12pt]{results/separate/mnist_separate/mnist_01234/input000\arabic{imagenumber}.png}
			}\forloop{imagenumber}{10}{\value{imagenumber} < 13}{
				& \includegraphics[width=14pt, height=12pt]{results/separate/mnist_separate/mnist_01234/input00\arabic{imagenumber}.png}
			} \\
			
			DVAE
			\forloop{imagenumber}{1}{\value{imagenumber} < 10}{
				& \includegraphics[width=14pt, height=13pt]{results/dvae/mnist/mnist_01234/frame000\arabic{imagenumber}.png}
			}\forloop{imagenumber}{10}{\value{imagenumber} < 13}{
				& \includegraphics[width=14pt, height=13pt]{results/dvae/mnist/mnist_01234/frame00\arabic{imagenumber}.png}
			}
			
		\end{tabular}
	\end{center}
	\caption{Synthesis of MNIST sequence 01234\dots}
	\label{fig:mnist}
\end{figure}
As an ablation study, we test our model on sequences of MNIST numbers. The aim is to see how well the DVAE captures the deterministic (predictable numbers) and stochastic (random writing style) aspects of the sequence, compared to en entirely linear system, and an autoregressive VAE model in which the VAE and VAR were learnt separately. \figref{fig:mnist} depicts the synthesis result for the number sequence 0123401234\dots. Our model captures the two essential features of this visual process. On the one hand, the particular number in a frame is entirely deterministic and can be inferred from the previous frame. On the other hand, the way the number is drawn is random and unpredictable. By contrast, the separate VAE+VAR model is only able to capture the appearance of the numbers and can not reproduce the number ordering, while the linear model generates hardly recognizable frames. 

\begin{figure}[h]
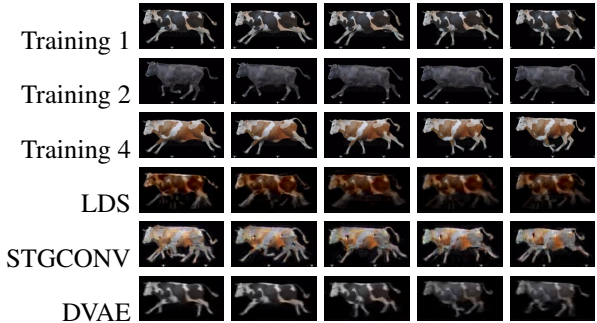

	\begin{center}
		\begin{tabular}{rccccccccccc}
			Training 1 
			\forloop[3]{imagenumber}{3}{\value{imagenumber} < 10}{
				& \includegraphics[width=32pt, height=17pt]{results/xie/cow/observed_sequence/sequence_1/00\arabic{imagenumber}.jpg}
			}\forloop[3]{imagenumber}{12}{\value{imagenumber} < 18}{
				& \includegraphics[width=32pt, height=17pt]{results/xie/cow/observed_sequence/sequence_1/0\arabic{imagenumber}.jpg}
			} \\
			Training 2
			\forloop[3]{imagenumber}{3}{\value{imagenumber} < 10}{
				& \includegraphics[width=32pt, height=17pt]{results/xie/cow/observed_sequence/sequence_2/00\arabic{imagenumber}.jpg}
			}\forloop[3]{imagenumber}{12}{\value{imagenumber} < 18}{
				& \includegraphics[width=32pt, height=17pt]{results/xie/cow/observed_sequence/sequence_2/0\arabic{imagenumber}.jpg}
			} \\
			Training 4
			\forloop[3]{imagenumber}{3}{\value{imagenumber} < 10}{
				& \includegraphics[width=32pt, height=17pt]{results/xie/cow/observed_sequence/sequence_4/00\arabic{imagenumber}.jpg}
			}\forloop[3]{imagenumber}{12}{\value{imagenumber} < 18}{
				& \includegraphics[width=32pt, height=17pt]{results/xie/cow/observed_sequence/sequence_4/0\arabic{imagenumber}.jpg}
			} \\
			LDS 
			\forloop[3]{imagenumber}{1}{\value{imagenumber} < 10}{
				& \includegraphics[width=32pt, height=17pt]{results/linear/cows/image000\arabic{imagenumber}.png}
			}\forloop[3]{imagenumber}{10}{\value{imagenumber} < 16}{
				& \includegraphics[width=32pt, height=17pt]{results/linear/cows/image00\arabic{imagenumber}.png}
			} \\
			
			STGCONV
			\forloop[3]{imagenumber}{1}{\value{imagenumber} < 10}{
				& \includegraphics[width=32pt, height=17pt]{results/xie/cow/image_000\arabic{imagenumber}.jpg}
			}\forloop[3]{imagenumber}{10}{\value{imagenumber} < 14}{
				& \includegraphics[width=32pt, height=17pt]{results/xie/cow/image_00\arabic{imagenumber}.jpg}
			} \\
			
			DVAE 
			\forloop[3]{imagenumber}{1}{\value{imagenumber} < 10}{
				& \includegraphics[width=32pt, height=17pt]{results/dvae/cows/frame000\arabic{imagenumber}.png}
			}\forloop[3]{imagenumber}{10}{\value{imagenumber} < 16}{
				& \includegraphics[width=32pt, height=17pt]{results/dvae/cows/frame00\arabic{imagenumber}.png}
			} \\
			
		\end{tabular}
	\end{center}
	\caption{Synthesis results for \emph{Running Cows}. Due to space constraints, two of the 5 training sequences are omitted.}
	\label{fig:cow}
\end{figure}
The \emph{Running Cows} sequences were used by the authors of \cite{Xie2017} to demonstrate the synthesis performance of STGCONV. \figref{fig:cow} depicts synthesis results for this sequence. We observe that although occasional discontinuities occur in the sequence synthesized by the DVAE, the overall running movement is accurately reproduced and unlike LDS or STGCONV, the DVAE does not introduce artifacts like additional legs. 
\begin{figure}
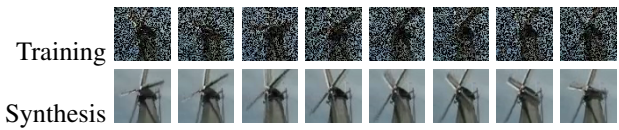

	\begin{center}
		\begin{tabular}{rcccccccc}
			Training 
			\forloop[2]{imagenumber}{1}{\value{imagenumber} < 10}{
				& \includegraphics[width=21pt, height=20pt]{results/original/incomplete/input000\arabic{imagenumber}.png}
			}\forloop[2]{imagenumber}{10}{\value{imagenumber} < 16}{
				& \includegraphics[width=21pt, height=20pt]{results/original/incomplete/input00\arabic{imagenumber}.png}
			} \\
			
			Synthesis 
			\forloop[2]{imagenumber}{1}{\value{imagenumber} < 10}{
				& \includegraphics[width=21pt, height=20pt]{results/dvae/incomplete/scene0000\arabic{imagenumber}.png}
			}\forloop[2]{imagenumber}{10}{\value{imagenumber} < 16}{
				& \includegraphics[width=21pt, height=20pt]{results/dvae/incomplete/scene000\arabic{imagenumber}.png}
			}
		\end{tabular}
	\end{center}
	\caption{Synthesizing a sequence learned on data obstructed by a salt+pepper mask.}
	\label{fig:incomplete}
\end{figure}
\begin{figure}
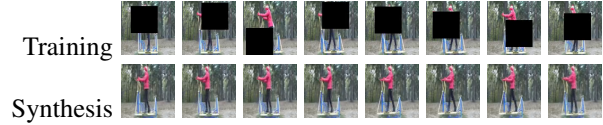

	\begin{center}
		\begin{tabular}{rcccccccccccccc}
			Training 
			\forloop[2]{imagenumber}{1}{\value{imagenumber} < 10}{
				& \includegraphics[width=20pt, height=20pt]{results/original/playing/frame000\arabic{imagenumber}.png}
			}\forloop[2]{imagenumber}{10}{\value{imagenumber} < 16}{
				& \includegraphics[width=20pt, height=20pt]{results/original/playing/frame00\arabic{imagenumber}.png}
			} \\
			
			Synthesis 
			\forloop[2]{imagenumber}{1}{\value{imagenumber} < 10}{
				& \includegraphics[width=20pt, height=20pt]{results/dvae/playing/frame000\arabic{imagenumber}.png}
			}\forloop[2]{imagenumber}{10}{\value{imagenumber} < 16}{
				& \includegraphics[width=20pt, height=20pt]{results/dvae/playing/frame00\arabic{imagenumber}.png}
			}
		\end{tabular}
	\end{center}
	\caption{Synthesizing a sequence learned on data obstructed by a rectangular mask.}
	\label{fig:incomplete2}
\end{figure}
Our model is also capable to learn from incomplete data, when the obstruction mask is given, albeit the results tend to become less steady over time. 
 \figref{fig:incomplete} depicts the synthesis of a video obstructed by 50\% salt+pepper noise. \figref{fig:incomplete2} depicts the synthesis of a video obstructed by 50\% rectangular mask. 

\subsection{Dynamic Textures}
\label{sub:dyntex}
\begin{figure}
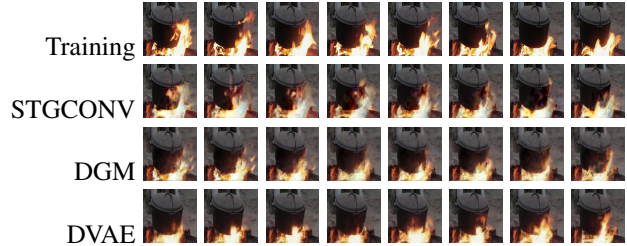

	\begin{center}
		\begin{tabular}{rcccccccccccccc}
			Training 
			\forloop{imagenumber}{7}{\value{imagenumber} < 10}{
				& \includegraphics[width=20pt, height=20pt]{results/original/stgconv_data/v11/image000\arabic{imagenumber}.png}
			}\forloop{imagenumber}{10}{\value{imagenumber} < 15}{
				& \includegraphics[width=20pt, height=20pt]{results/original/stgconv_data/v11/image00\arabic{imagenumber}.png}
			} \\
			
			
			STGCONV 
			\forloop{imagenumber}{7}{\value{imagenumber} < 10}{
				& \includegraphics[width=20pt, height=20pt]{results/xie/v11/image_000\arabic{imagenumber}.jpg}
			}\forloop{imagenumber}{10}{\value{imagenumber} < 15}{
				& \includegraphics[width=20pt, height=20pt]{results/xie/v11/image_00\arabic{imagenumber}.jpg}
			} \\
			
			DGM 
			\forloop{imagenumber}{7}{\value{imagenumber} < 10}{
				& \includegraphics[width=20pt, height=20pt]{results/xie2/v11/00\arabic{imagenumber}.png}
			}\forloop{imagenumber}{10}{\value{imagenumber} < 15}{
				& \includegraphics[width=20pt, height=20pt]{results/xie2/v11/0\arabic{imagenumber}.png}
			} \\

			DVAE 
			\forloop{imagenumber}{6}{\value{imagenumber} < 10}{
				& \includegraphics[width=20pt, height=20pt]{results/dvae/stgconv/v11/scene000\arabic{imagenumber}.png}
			}\forloop{imagenumber}{10}{\value{imagenumber} < 14}{
				& \includegraphics[width=20pt, height=20pt]{results/dvae/stgconv/v11/scene00\arabic{imagenumber}.png}
			}
		\end{tabular}
	\end{center}
	\caption{Synthesis of the dynamic texture \emph{Fire Pot}}
	\label{fig:firepot}
\end{figure}
We synthesize sequences of eleven dynamic textures that were provided in the supplementary material of \cite{Xie2017}. Table~\ref{tab:fid} summarizes the resulting FID scores.
As a demonstration, \figref{fig:firepot} depicts the synthesis results for the \emph{Fire Pot} video. Compared to the STGCONV method, the DVAE yields slightly blurrier results. However, this advantage of the STGCONV can be explained by its tendency to reproduce the training sequence. This behavior can be observed in \figref{fig:firepot}. Comparing the training sequence to the STGCONV result indicates that, at each time step, the STGCONV frames are a slightly perturbed version of the training frames, whereas DVAE produces frames that evolve differently over time, while maintaining a natural dynamic. The DGM produces less predictable sequences, but the frame transition appears less natural, resembling a fading of one frame into the other.

\begin{table}
	\begin{center}
		\begin{tabular}{ccccc}
			\textbf{Sequence} & \textbf{LDS} &\ \textbf{STGCONV}&\ \textbf{DGM} &\ \textbf{DVAE} \\
			\hline
			Cows & 267.0 & 311.2 & - &\textbf{150.4} \\
			Flowing Water & 221.0 & - & - &  \textbf{163.9} \\
			Boiling Water & \textbf{154.7} & - & - &  175.8 \\
			Sea & 108.9 & - & - &  \textbf{64.1} \\
			River & 238.4 & \textbf{103.3} & - &110.1 \\
			Mountain Stream & 224.9 & - & - &  \textbf{216.2} \\
			Spring Water  & 333.4 & \textbf{233.3} & - &235.9 \\
			Fountain  & 241.9 & 271.9 & - &\textbf{135.8} \\
			Waterfall  & 381.4 & \textbf{236.7} & - &336.1 \\
			Washing Machine & \textbf{115.1} & - & - & 128.9 \\			
			Flashing Lights  & 191.0 & 166.7 & 257.8 &\textbf{128.3} \\
			Fire Pot  & 189.6 & 172.2 & 146.3 &\textbf{124.9} \\ 
		\end{tabular}
	\end{center}
	\caption{FID Scores}
	\label{tab:fid}
\end{table}

\section{Conclusion}
We proposed a deep generative model for visual processes based on an auto-regressive state space model and a Variational Autoencoder. Despite being based on a dynamic behavior of a mathematically very simple form, our model is capable to reproduce various kinds of highly non-linear and non-Gaussian visual processes and compete with state-of-the-art approaches for dynamic texture synthesis.

\bibliographystyle{IEEEbib}
\bibliography{refs}

\end{document}


\title{Appendix To Submission \# 371}

\author{First Author \\
Institution1\\
{\tt\small firstauthor@i1.org}
\and
Second Author \\
Institution2\\
{\tt\small secondauthor@i2.org}
}

\maketitle
\ifwacvfinal\thispagestyle{empty}\fi


\newtheorem{mydef}{Definition}
\newtheorem{myasm}{Assumption}
\newtheorem{myrem}{Remark}
\newtheorem{proposition}{Proposition}
\newcommand{\eor}{\hfill$\lozenge$}
\newcommand{\eop}{\hfill$\square$}


	\begin{appendices}
\section{Generalization to Higher-order Markov Processes}
\label{sub:higher_markov}
The proposed method is designed for first-order Markov processes only. However, the presented approach can be theoretically generalized to Markov processes of order $m\geq 1$. We present a procedure that could be employed for this purpose. Further investigation is necessary to evaluate the feasibility and practical applicability of such an approach.

The general procedure is as follows. 
\begin{enumerate}
	\item Set up an appropriate block matrix model for the joint probability of $m+1$ succeeding observations.
	\item Build a dynamic layer that performs a multiplication with a lower-triangular $(m+1)\times (m+1)$ block matrix $\mF$.
	\item Introduce regularizers to preserve the block Toeplitz structure of the covariance matrix,
	\item Compute the VAR parameters from the resulting covariance matrix.
\end{enumerate}
We illustrate the procedure for the case $m=2$.

\begin{enumerate}
	\item Due to the stationarity assumption, the covariance matrix for three succeeding observations $\rvh_t,\rvh_{t+1},\rvh_{t+2}$ has the form
	\begin{equation}
 \mSigma_2=\begin{bmatrix}
	\mI_n & \Cov(\rvh_t,\rvh_{t+1}) & \Cov(\rvh_t,\rvh_{t+2})\\
	\Cov(\rvh_t,\rvh_{t+1})^\top & \mI_n & \Cov(\rvh_t,\rvh_{t+1})\\
	\Cov(\rvh_t,\rvh_{t+2})^\top & \Cov(\rvh_t,\rvh_{t+1})^\top & \mI_n
	\end{bmatrix}.
	\end{equation}
	\item The matrix describing the dynamic layer has the form
	\begin{equation}
	\mF=\begin{bmatrix}
	\mI_n & & \\
	\mF_1 & \mF_2 & \\
	\mF_3 & \mF_4 & \mF_5.
	\end{bmatrix}
	\end{equation}
	The outputs of the dynamic layer will thus have the distribution
	\begin{equation}
	p\left(\begin{bmatrix}\rvh_1^\top & \rvh_2^\top & \rvh_3^\top\end{bmatrix}^\top\right)=\mathcal{N}(\begin{bmatrix}\rvh_1^\top & \rvh_2^\top & \rvh_3^\top\end{bmatrix}^\top;\ 0,\mF\mF^\top).
	\end{equation}
	We thus need to achieve
	\begin{equation}
	\mSigma_2\approx \mF\mF^\top=\begin{bmatrix}
	\mI_n & \mF_1^\top & \mF_3^\top \\
	\mF_1 & \mF_1\mF_1^\top+\mF_2\mF_2^\top &  \mF_1\mF_3^\top+\mF_2\mF_4^\top \\
	\mF_3^\top & \mF_3\mF_1^\top +\mF_4\mF_2^\top & \mF_3\mF_3^\top+\mF_4\mF_4^\top+\mF_5\mF_5^\top
	\end{bmatrix}.
	\end{equation}
	\item This can be ensured by using the regularizer
	\begin{equation}
	\begin{split}
	R(\mF)=\lambda_1\|\mF_1\mF_1^\top+\mF_2\mF_2^\top-\mI_n\|_F^2+\lambda_2\|\mF_3\mF_3^\top+\mF_4\mF_4^\top+\mF_5\mF_5^\top-\mI_n\|_F^2\\+\lambda_3\|\mF_3\mF_1^\top+\mF_4\mF_2^\top-\mF_1\|_F^2,
	\end{split}
	\end{equation}
	with $\lambda_1,\lambda_2,\lambda_3>0$.
	\item A second-order VAR model has the form
	\begin{eqnarray}
	\vh_{t+2}=\mA_0\vh_{t}+\mA_1\vh_{t+1}+\mB\vv_t,\;\rvv_t\sim\mathcal N(\rvv_t,0,\mI_n).
	\end{eqnarray}
	By our assumptions on the process, this yields the system of equations,
	\begin{equation}
	\begin{split}
	\Cov(\rvh_{t},\rvh_{t})&=\mA_0\mA_0^\top+\mA_1\mA_1^\top+\mA_0\Cov(\rvh_t,\rvh_{t+1})\mA_1^\top\\
	&\; \; \; +\mA_1\Cov(\rvh_{t},\rvh_{t+1})^\top\mA_0^\top+\mB\mB^\top\\
	\Cov(\rvh_{t},\rvh_{t+1})&=\Cov(\rvh_t,\rvh_{t+1})^\top\mA_0^\top+\mA_1^\top,\\
	\Cov(\rvh_t,\rvh_{t+2})&=\mA_0^\top + \Cov(\rvh_t,\rvh_{t+1})\mA_1^\top.
	\end{split}
	\end{equation}
	Thus, we can infer $\mA_0, \mA_1$ and $\mB$ via solving
	\begin{equation}
	\begin{split}
	\mF_1^\top&=\mF_1\mA_0^\top+\mA_1^\top,\\
	\mF_3&=\mA_0 + \mA_1\mF_1,\\
	\mI_n&=\mA_0\mA_0^\top+\mA_1\mA_1^\top+\mA_0\mF_1^\top\mA_1^\top+\mA_1\mF_1\mA_0^\top+\mB\mB^\top.
	\end{split}
	\end{equation}
\end{enumerate}		
\clearpage
\section{Additional Experimental Analysis}
\newcounter{imagenumber}
\newcounter{mnistnumber}
\setlength{\tabcolsep}{0.3pt}

\subsection{MNIST}
\figref{fig:mnist12345} - \figref{fig:mnist56789} depict additional synthesis results of the MNIST experiment.

\forloop[11111]{mnistnumber}{12345}{\value{mnistnumber} < 60000}{
	\begin{figure}[h]
		\begin{center}
			\begin{tabular}{rcccccccccccccccccccc}	
				Training 
				\forloop{imagenumber}{1}{\value{imagenumber} < 10}{
					& \includegraphics[width=16pt, height=16pt]{results/original/mnist/mnist_\arabic{mnistnumber}/image000\arabic{imagenumber}.png}
				}\forloop{imagenumber}{10}{\value{imagenumber} < 21}{
					& \includegraphics[width=16pt, height=16pt]{results/original/mnist/mnist_\arabic{mnistnumber}/image00\arabic{imagenumber}.png}
				} \\
				
				LDS 
				\forloop{imagenumber}{1}{\value{imagenumber} < 10}{
					& \includegraphics[width=16pt, height=16pt]{results/linear/mnist_linear/mnist_\arabic{mnistnumber}/image000\arabic{imagenumber}.png}
				}\forloop{imagenumber}{10}{\value{imagenumber} < 21}{
					& \includegraphics[width=16pt, height=16pt]{results/linear/mnist_linear/mnist_\arabic{mnistnumber}/image00\arabic{imagenumber}.png}
				} \\
				
				VAE+VAR
				\forloop{imagenumber}{1}{\value{imagenumber} < 10}{
					& \includegraphics[width=16pt, height=16pt]{results/separate/mnist_separate/mnist_\arabic{mnistnumber}/input000\arabic{imagenumber}.png}
				}\forloop{imagenumber}{10}{\value{imagenumber} < 21}{
					& \includegraphics[width=16pt, height=16pt]{results/separate/mnist_separate/mnist_\arabic{mnistnumber}/input00\arabic{imagenumber}.png}
				} \\
				
				DVAE
				\forloop{imagenumber}{1}{\value{imagenumber} < 10}{
					& \includegraphics[width=16pt, height=16pt]{results/dvae/mnist/mnist_\arabic{mnistnumber}/frame000\arabic{imagenumber}.png}
				}\forloop{imagenumber}{10}{\value{imagenumber} < 21}{
					& \includegraphics[width=16pt, height=16pt]{results/dvae/mnist/mnist_\arabic{mnistnumber}/frame00\arabic{imagenumber}.png}
				}
				
			\end{tabular}
		\end{center}
		\caption{Synthesis of MNIST sequence \arabic{mnistnumber}\arabic{mnistnumber}\dots}
		\label{fig:mnist\arabic{mnistnumber}}
	\end{figure}
}

\clearpage
\subsection{Small NORB}
\figref{fig:norb4} - \figref{fig:norb8} depict additional synthesis results of the Small NORB experiment.
\begin{figure}[h]
	\begin{center}
		\begin{tabular}{rccccccccccccccccc}
			Training 
			\forloop{imagenumber}{1}{\value{imagenumber} < 10}{
				& \includegraphics[width=20pt, height=20pt]{results/original/norb/norb_cat_0_inst_4/image000\arabic{imagenumber}.png}
			}\forloop{imagenumber}{10}{\value{imagenumber} < 18}{
				& \includegraphics[width=20pt, height=20pt]{results/original/norb/norb_cat_0_inst_4/image00\arabic{imagenumber}.png}
			} \\
			
			LDS 
			\forloop{imagenumber}{1}{\value{imagenumber} < 10}{
				& \includegraphics[width=20pt, height=20pt]{results/linear/norb_linear/norb_cat_0_inst_4/image000\arabic{imagenumber}.png}
			}\forloop{imagenumber}{10}{\value{imagenumber} < 18}{
				& \includegraphics[width=20pt, height=20pt]{results/linear/norb_linear/norb_cat_0_inst_4/image00\arabic{imagenumber}.png}
			} \\
			
			VAE+VAR
			\forloop{imagenumber}{1}{\value{imagenumber} < 10}{
				& \includegraphics[width=20pt, height=20pt]{results/separate/norb_separate/norb_cat_0_inst_4/input000\arabic{imagenumber}.png}
			}\forloop{imagenumber}{10}{\value{imagenumber} < 18}{
				& \includegraphics[width=20pt, height=20pt]{results/separate/norb_separate/norb_cat_0_inst_4/input00\arabic{imagenumber}.png}
			} \\
			
			DVAE
			\forloop{imagenumber}{1}{\value{imagenumber} < 10}{
				& \includegraphics[width=20pt, height=20pt]{results/dvae/norb/norb_cat_0_inst_4/image000\arabic{imagenumber}.png}
			}\forloop{imagenumber}{10}{\value{imagenumber} < 18}{
				& \includegraphics[width=20pt, height=20pt]{results/dvae/norb/norb_cat_0_inst_4/image00\arabic{imagenumber}.png}
			}
			
		\end{tabular}
	\end{center}
	\caption{Synthesis of a rotation sequence of Small NORB images (Category 0, Instance 4)}
	\label{fig:norb4}
\end{figure}
\newcounter{norbnumber}
\forloop{norbnumber}{6}{\value{norbnumber} < 10}{
	\begin{figure}[h]
		\begin{center}
			\begin{tabular}{rccccccccccccccccc}
				Training 
				\forloop{imagenumber}{1}{\value{imagenumber} < 10}{
					& \includegraphics[width=20pt, height=20pt]{results/original/norb/norb_cat_0_inst_\arabic{norbnumber}/image000\arabic{imagenumber}.png}
				}\forloop{imagenumber}{10}{\value{imagenumber} < 18}{
					& \includegraphics[width=20pt, height=20pt]{results/original/norb/norb_cat_0_inst_\arabic{norbnumber}/image00\arabic{imagenumber}.png}
				} \\
				
				LDS 
				\forloop{imagenumber}{1}{\value{imagenumber} < 10}{
					& \includegraphics[width=20pt, height=20pt]{results/linear/norb_linear/norb_cat_0_inst_\arabic{norbnumber}/image000\arabic{imagenumber}.png}
				}\forloop{imagenumber}{10}{\value{imagenumber} < 18}{
					& \includegraphics[width=20pt, height=20pt]{results/linear/norb_linear/norb_cat_0_inst_\arabic{norbnumber}/image00\arabic{imagenumber}.png}
				} \\
				
				VAE+VAR
				\forloop{imagenumber}{1}{\value{imagenumber} < 10}{
					& \includegraphics[width=20pt, height=20pt]{results/separate/norb_separate/norb_cat_0_inst_\arabic{norbnumber}/input000\arabic{imagenumber}.png}
				}\forloop{imagenumber}{10}{\value{imagenumber} < 18}{
					& \includegraphics[width=20pt, height=20pt]{results/separate/norb_separate/norb_cat_0_inst_\arabic{norbnumber}/input00\arabic{imagenumber}.png}
				} \\
				
				DVAE
				\forloop{imagenumber}{1}{\value{imagenumber} < 10}{
					& \includegraphics[width=20pt, height=20pt]{results/dvae/norb/norb_cat_0_inst_\arabic{norbnumber}/image000\arabic{imagenumber}.png}
				}\forloop{imagenumber}{10}{\value{imagenumber} < 18}{
					& \includegraphics[width=20pt, height=20pt]{results/dvae/norb/norb_cat_0_inst_\arabic{norbnumber}/image00\arabic{imagenumber}.png}
				}
				
			\end{tabular}
		\end{center}
		\caption{Synthesis of a rotation sequence of Small NORB images (Category 0, Instance \arabic{norbnumber})}
		\label{fig:norb\arabic{norbnumber}}
	\end{figure}
}
\clearpage
\subsection{Dynamic Textures}
\label{sub:dyntex}
\figref{fig:flwater} - \figref{fig:firepot} depict additional synthesis results of dynamic textures. Note that the STGCONV frames were synthesized in a slightly higher resolution, $224\times 224$.

\begin{figure}[h]
	\begin{center}
		\begin{tabular}{rcccccccccccccc}
			Training 
			\forloop{imagenumber}{2}{\value{imagenumber} < 10}{
				& \includegraphics[width=28pt, height=28pt]{results/original/stgconv_data/v0/image000\arabic{imagenumber}.png}
			}\forloop{imagenumber}{10}{\value{imagenumber} < 15}{
				& \includegraphics[width=28pt, height=28pt]{results/original/stgconv_data/v0/image00\arabic{imagenumber}.png}
			} \\
			
			LDS 
			\forloop{imagenumber}{2}{\value{imagenumber} < 10}{
				& \includegraphics[width=28pt, height=28pt]{results/linear/stgconv_linear/v0/image000\arabic{imagenumber}.png}
			}\forloop{imagenumber}{10}{\value{imagenumber} < 15}{
				& \includegraphics[width=28pt, height=28pt]{results/linear/stgconv_linear/v0/image00\arabic{imagenumber}.png}
			} \\
			
			DVAE 
			\forloop{imagenumber}{1}{\value{imagenumber} < 10}{
				& \includegraphics[width=28pt, height=28pt]{results/dvae/stgconv/v0/scene000\arabic{imagenumber}.png}
			}\forloop{imagenumber}{10}{\value{imagenumber} < 14}{
				& \includegraphics[width=28pt, height=28pt]{results/dvae/stgconv/v0/scene00\arabic{imagenumber}.png}
			}
		\end{tabular}
	\end{center}
	\caption{Synthesis of dynamic texture \emph{Flowing Water}}
	\label{fig:flwater}
\end{figure}

\begin{figure}[h]
	\begin{center}
		\begin{tabular}{rcccccccccccccc}
			Training 
			\forloop{imagenumber}{2}{\value{imagenumber} < 10}{
				& \includegraphics[width=28pt, height=28pt]{results/original/stgconv_data/v1/image000\arabic{imagenumber}.png}
			}\forloop{imagenumber}{10}{\value{imagenumber} < 15}{
				& \includegraphics[width=28pt, height=28pt]{results/original/stgconv_data/v1/image00\arabic{imagenumber}.png}
			} \\
			
			LDS 
			\forloop{imagenumber}{2}{\value{imagenumber} < 10}{
				& \includegraphics[width=28pt, height=28pt]{results/linear/stgconv_linear/v1/image000\arabic{imagenumber}.png}
			}\forloop{imagenumber}{10}{\value{imagenumber} < 15}{
				& \includegraphics[width=28pt, height=28pt]{results/linear/stgconv_linear/v1/image00\arabic{imagenumber}.png}
			} \\
			
			DVAE 
			\forloop{imagenumber}{1}{\value{imagenumber} < 10}{
				& \includegraphics[width=28pt, height=28pt]{results/dvae/stgconv/v1/scene000\arabic{imagenumber}.png}
			}\forloop{imagenumber}{10}{\value{imagenumber} < 14}{
				& \includegraphics[width=28pt, height=28pt]{results/dvae/stgconv/v1/scene00\arabic{imagenumber}.png}
			}
		\end{tabular}
	\end{center}
	\caption{Synthesis of dynamic texture \emph{Boiling Water}}
	\label{fig:blwater}
\end{figure}

\begin{figure}[h]
	\begin{center}
		\begin{tabular}{rcccccccccccccc}
			Training 
			\forloop{imagenumber}{2}{\value{imagenumber} < 10}{
				& \includegraphics[width=28pt, height=28pt]{results/original/stgconv_data/v2/image000\arabic{imagenumber}.png}
			}\forloop{imagenumber}{10}{\value{imagenumber} < 15}{
				& \includegraphics[width=28pt, height=28pt]{results/original/stgconv_data/v2/image00\arabic{imagenumber}.png}
			} \\
			
			LDS 
			\forloop{imagenumber}{2}{\value{imagenumber} < 10}{
				& \includegraphics[width=28pt, height=28pt]{results/linear/stgconv_linear/v2/image000\arabic{imagenumber}.png}
			}\forloop{imagenumber}{10}{\value{imagenumber} < 15}{
				& \includegraphics[width=28pt, height=28pt]{results/linear/stgconv_linear/v2/image00\arabic{imagenumber}.png}
			} \\
			
			DVAE 
			\forloop{imagenumber}{1}{\value{imagenumber} < 10}{
				& \includegraphics[width=28pt, height=28pt]{results/dvae/stgconv/v2/scene000\arabic{imagenumber}.png}
			}\forloop{imagenumber}{10}{\value{imagenumber} < 14}{
				& \includegraphics[width=28pt, height=28pt]{results/dvae/stgconv/v2/scene00\arabic{imagenumber}.png}
			}
		\end{tabular}
	\end{center}
	\caption{Synthesis of dynamic texture \emph{Sea}}
	\label{fig:sea}
\end{figure}

\begin{figure}[h]
	\begin{center}
		\begin{tabular}{rcccccccccccccc}
			Training 
			\forloop{imagenumber}{2}{\value{imagenumber} < 10}{
				& \includegraphics[width=28pt, height=28pt]{results/original/stgconv_data/v3/image000\arabic{imagenumber}.png}
			}\forloop{imagenumber}{10}{\value{imagenumber} < 15}{
				& \includegraphics[width=28pt, height=28pt]{results/original/stgconv_data/v3/image00\arabic{imagenumber}.png}
			} \\
			
			LDS 
			\forloop{imagenumber}{2}{\value{imagenumber} < 10}{
				& \includegraphics[width=28pt, height=28pt]{results/linear/stgconv_linear/v3/image000\arabic{imagenumber}.png}
			}\forloop{imagenumber}{10}{\value{imagenumber} < 15}{
				& \includegraphics[width=28pt, height=28pt]{results/linear/stgconv_linear/v3/image00\arabic{imagenumber}.png}
			} \\
			
			STGCONV 
			\forloop{imagenumber}{2}{\value{imagenumber} < 10}{
				& \includegraphics[width=28pt, height=28pt]{results/xie/v3/image_000\arabic{imagenumber}.jpg}
			}\forloop{imagenumber}{10}{\value{imagenumber} < 15}{
				& \includegraphics[width=28pt, height=28pt]{results/xie/v3/image_00\arabic{imagenumber}.jpg}
			} \\
			
			DVAE 
			\forloop{imagenumber}{1}{\value{imagenumber} < 10}{
				& \includegraphics[width=28pt, height=28pt]{results/dvae/stgconv/v3/scene000\arabic{imagenumber}.png}
			}\forloop{imagenumber}{10}{\value{imagenumber} < 14}{
				& \includegraphics[width=28pt, height=28pt]{results/dvae/stgconv/v3/scene00\arabic{imagenumber}.png}
			}
		\end{tabular}
	\end{center}
	\caption{Synthesis of dynamic texture \emph{River}}
	\label{fig:river}
\end{figure}

\begin{figure}[h]
	\begin{center}
		\begin{tabular}{rcccccccccccccc}
			Training 
			\forloop{imagenumber}{2}{\value{imagenumber} < 10}{
				& \includegraphics[width=28pt, height=28pt]{results/original/stgconv_data/v5/image000\arabic{imagenumber}.png}
			}\forloop{imagenumber}{10}{\value{imagenumber} < 15}{
				& \includegraphics[width=28pt, height=28pt]{results/original/stgconv_data/v5/image00\arabic{imagenumber}.png}
			} \\
			
			LDS 
			\forloop{imagenumber}{2}{\value{imagenumber} < 10}{
				& \includegraphics[width=28pt, height=28pt]{results/linear/stgconv_linear/v5/image000\arabic{imagenumber}.png}
			}\forloop{imagenumber}{10}{\value{imagenumber} < 15}{
				& \includegraphics[width=28pt, height=28pt]{results/linear/stgconv_linear/v5/image00\arabic{imagenumber}.png}
			} \\
			
			STGCONV 
			\forloop{imagenumber}{2}{\value{imagenumber} < 10}{
				& \includegraphics[width=28pt, height=28pt]{results/xie/v5/image_000\arabic{imagenumber}.jpg}
			}\forloop{imagenumber}{10}{\value{imagenumber} < 15}{
				& \includegraphics[width=28pt, height=28pt]{results/xie/v5/image_00\arabic{imagenumber}.jpg}
			} \\
			
			DVAE 
			\forloop{imagenumber}{1}{\value{imagenumber} < 10}{
				& \includegraphics[width=28pt, height=28pt]{results/dvae/stgconv/v5/scene000\arabic{imagenumber}.png}
			}\forloop{imagenumber}{10}{\value{imagenumber} < 14}{
				& \includegraphics[width=28pt, height=28pt]{results/dvae/stgconv/v5/scene00\arabic{imagenumber}.png}
			}
		\end{tabular}
	\end{center}
	\caption{Synthesis of dynamic texture \emph{Spring water}}
\end{figure}

\begin{figure}[h]
	\begin{center}
		\begin{tabular}{rcccccccccccccc}
			Training 
			\forloop{imagenumber}{2}{\value{imagenumber} < 10}{
				& \includegraphics[width=28pt, height=28pt]{results/original/stgconv_data/v4/image000\arabic{imagenumber}.png}
			}\forloop{imagenumber}{10}{\value{imagenumber} < 15}{
				& \includegraphics[width=28pt, height=28pt]{results/original/stgconv_data/v4/image00\arabic{imagenumber}.png}
			} \\
			
			LDS 
			\forloop{imagenumber}{2}{\value{imagenumber} < 10}{
				& \includegraphics[width=28pt, height=28pt]{results/linear/stgconv_linear/v4/image000\arabic{imagenumber}.png}
			}\forloop{imagenumber}{10}{\value{imagenumber} < 15}{
				& \includegraphics[width=28pt, height=28pt]{results/linear/stgconv_linear/v4/image00\arabic{imagenumber}.png}
			} \\
			
			DVAE 
			\forloop{imagenumber}{1}{\value{imagenumber} < 10}{
				& \includegraphics[width=28pt, height=28pt]{results/dvae/stgconv/v4/scene000\arabic{imagenumber}.png}
			}\forloop{imagenumber}{10}{\value{imagenumber} < 14}{
				& \includegraphics[width=28pt, height=28pt]{results/dvae/stgconv/v4/scene00\arabic{imagenumber}.png}
			}
		\end{tabular}
	\end{center}
	\caption{Synthesis of dynamic texture \emph{Mountain Stream}}
	\label{fig:mountainstream}
\end{figure}

\begin{figure}[h]
	\begin{center}
		\begin{tabular}{rcccccccccccccc}
			Training 
			\forloop{imagenumber}{2}{\value{imagenumber} < 10}{
				& \includegraphics[width=28pt, height=28pt]{results/original/stgconv_data/v6/image000\arabic{imagenumber}.png}
			}\forloop{imagenumber}{10}{\value{imagenumber} < 15}{
				& \includegraphics[width=28pt, height=28pt]{results/original/stgconv_data/v6/image00\arabic{imagenumber}.png}
			} \\
			
			LDS 
			\forloop{imagenumber}{2}{\value{imagenumber} < 10}{
				& \includegraphics[width=28pt, height=28pt]{results/linear/stgconv_linear/v6/image000\arabic{imagenumber}.png}
			}\forloop{imagenumber}{10}{\value{imagenumber} < 15}{
				& \includegraphics[width=28pt, height=28pt]{results/linear/stgconv_linear/v6/image00\arabic{imagenumber}.png}
			} \\
			
			STGCONV 
			\forloop{imagenumber}{2}{\value{imagenumber} < 10}{
				& \includegraphics[width=28pt, height=28pt]{results/xie/v6/image_000\arabic{imagenumber}.jpg}
			}\forloop{imagenumber}{10}{\value{imagenumber} < 15}{
				& \includegraphics[width=28pt, height=28pt]{results/xie/v6/image_00\arabic{imagenumber}.jpg}
			} \\
			
			DVAE 
			\forloop{imagenumber}{1}{\value{imagenumber} < 10}{
				& \includegraphics[width=28pt, height=28pt]{results/dvae/stgconv/v6/scene000\arabic{imagenumber}.png}
			}\forloop{imagenumber}{10}{\value{imagenumber} < 14}{
				& \includegraphics[width=28pt, height=28pt]{results/dvae/stgconv/v6/scene00\arabic{imagenumber}.png}
			}
		\end{tabular}
	\end{center}
	\caption{Synthesis of dynamic texture \emph{Fountain}}
\end{figure}

\begin{figure}[h]
	\begin{center}
		\begin{tabular}{rcccccccccccccc}
			Training 
			\forloop{imagenumber}{2}{\value{imagenumber} < 10}{
				& \includegraphics[width=28pt, height=28pt]{results/original/stgconv_data/v7/image000\arabic{imagenumber}.png}
			}\forloop{imagenumber}{10}{\value{imagenumber} < 15}{
				& \includegraphics[width=28pt, height=28pt]{results/original/stgconv_data/v7/image00\arabic{imagenumber}.png}
			} \\
			
			LDS 
			\forloop{imagenumber}{2}{\value{imagenumber} < 10}{
				& \includegraphics[width=28pt, height=28pt]{results/linear/stgconv_linear/v7/image000\arabic{imagenumber}.png}
			}\forloop{imagenumber}{10}{\value{imagenumber} < 15}{
				& \includegraphics[width=28pt, height=28pt]{results/linear/stgconv_linear/v7/image00\arabic{imagenumber}.png}
			} \\
			
			STGCONV 
			\forloop{imagenumber}{2}{\value{imagenumber} < 10}{
				& \includegraphics[width=28pt, height=28pt]{results/xie/v7/image_000\arabic{imagenumber}.jpg}
			}\forloop{imagenumber}{10}{\value{imagenumber} < 15}{
				& \includegraphics[width=28pt, height=28pt]{results/xie/v7/image_00\arabic{imagenumber}.jpg}
			} \\
			
			DVAE 
			\forloop{imagenumber}{1}{\value{imagenumber} < 10}{
				& \includegraphics[width=28pt, height=28pt]{results/dvae/stgconv/v7/scene000\arabic{imagenumber}.png}
			}\forloop{imagenumber}{10}{\value{imagenumber} < 14}{
				& \includegraphics[width=28pt, height=28pt]{results/dvae/stgconv/v7/scene00\arabic{imagenumber}.png}
			}
		\end{tabular}
	\end{center}
	\caption{Synthesis of dynamic texture \emph{Waterfall}}
\end{figure}

\begin{figure}[h]
	\begin{center}
		\begin{tabular}{rcccccccccccccc}
			Training 
			\forloop{imagenumber}{2}{\value{imagenumber} < 10}{
				& \includegraphics[width=28pt, height=28pt]{results/original/stgconv_data/v8/image000\arabic{imagenumber}.png}
			}\forloop{imagenumber}{10}{\value{imagenumber} < 15}{
				& \includegraphics[width=28pt, height=28pt]{results/original/stgconv_data/v8/image00\arabic{imagenumber}.png}
			} \\
			
			LDS 
			\forloop{imagenumber}{2}{\value{imagenumber} < 10}{
				& \includegraphics[width=28pt, height=28pt]{results/linear/stgconv_linear/v8/image000\arabic{imagenumber}.png}
			}\forloop{imagenumber}{10}{\value{imagenumber} < 15}{
				& \includegraphics[width=28pt, height=28pt]{results/linear/stgconv_linear/v8/image00\arabic{imagenumber}.png}
			} \\
			
			DVAE 
			\forloop{imagenumber}{1}{\value{imagenumber} < 10}{
				& \includegraphics[width=28pt, height=28pt]{results/dvae/stgconv/v8/scene000\arabic{imagenumber}.png}
			}\forloop{imagenumber}{10}{\value{imagenumber} < 14}{
				& \includegraphics[width=28pt, height=28pt]{results/dvae/stgconv/v8/scene00\arabic{imagenumber}.png}
			}
		\end{tabular}
	\end{center}
	\caption{Synthesis of dynamic texture \emph{Washing Machine}}
	\label{fig:washing}
\end{figure}

\begin{figure}[h]
	\begin{center}
		\begin{tabular}{rcccccccccccccc}
			Training 
			\forloop{imagenumber}{2}{\value{imagenumber} < 10}{
				& \includegraphics[width=28pt, height=28pt]{results/original/stgconv_data/v9/image000\arabic{imagenumber}.png}
			}\forloop{imagenumber}{10}{\value{imagenumber} < 15}{
				& \includegraphics[width=28pt, height=28pt]{results/original/stgconv_data/v9/image00\arabic{imagenumber}.png}
			} \\
			
			LDS 
			\forloop{imagenumber}{2}{\value{imagenumber} < 10}{
				& \includegraphics[width=28pt, height=28pt]{results/linear/stgconv_linear/v9/image000\arabic{imagenumber}.png}
			}\forloop{imagenumber}{10}{\value{imagenumber} < 15}{
				& \includegraphics[width=28pt, height=28pt]{results/linear/stgconv_linear/v9/image00\arabic{imagenumber}.png}
			} \\
			
			STGCONV 
			\forloop{imagenumber}{2}{\value{imagenumber} < 10}{
				& \includegraphics[width=28pt, height=28pt]{results/xie/v9/image_000\arabic{imagenumber}.jpg}
			}\forloop{imagenumber}{10}{\value{imagenumber} < 15}{
				& \includegraphics[width=28pt, height=28pt]{results/xie/v9/image_00\arabic{imagenumber}.jpg}
			} \\

			DGM 
			\forloop{imagenumber}{1}{\value{imagenumber} < 10}{
				& \includegraphics[width=28pt, height=28pt]{results/xie2/v9/00\arabic{imagenumber}.png}
			}\forloop{imagenumber}{10}{\value{imagenumber} < 14}{
				& \includegraphics[width=28pt, height=28pt]{results/xie2/v9/0\arabic{imagenumber}.png}
			} \\

			DVAE 
			\forloop{imagenumber}{1}{\value{imagenumber} < 10}{
				& \includegraphics[width=28pt, height=28pt]{results/dvae/stgconv/v9/scene000\arabic{imagenumber}.png}
			}\forloop{imagenumber}{10}{\value{imagenumber} < 14}{
				& \includegraphics[width=28pt, height=28pt]{results/dvae/stgconv/v9/scene00\arabic{imagenumber}.png}
			}
		\end{tabular}
	\end{center}
	\caption{Synthesis of dynamic texture \emph{Flashing lights}}
	\label{fig:lights}
\end{figure}

\begin{figure}[h]
	\begin{center}
		\begin{tabular}{rcccccccccccccc}
			Training 
			\forloop{imagenumber}{2}{\value{imagenumber} < 10}{
				& \includegraphics[width=28pt, height=28pt]{results/original/stgconv_data/v11/image000\arabic{imagenumber}.png}
			}\forloop{imagenumber}{10}{\value{imagenumber} < 15}{
				& \includegraphics[width=28pt, height=28pt]{results/original/stgconv_data/v11/image00\arabic{imagenumber}.png}
			} \\
			
			LDS 
			\forloop{imagenumber}{2}{\value{imagenumber} < 10}{
				& \includegraphics[width=28pt, height=28pt]{results/linear/stgconv_linear/v11/image000\arabic{imagenumber}.png}
			}\forloop{imagenumber}{10}{\value{imagenumber} < 15}{
				& \includegraphics[width=28pt, height=28pt]{results/linear/stgconv_linear/v11/image00\arabic{imagenumber}.png}
			} \\
			
			STGCONV 
			\forloop{imagenumber}{2}{\value{imagenumber} < 10}{
				& \includegraphics[width=28pt, height=28pt]{results/xie/v11/image_000\arabic{imagenumber}.jpg}
			}\forloop{imagenumber}{10}{\value{imagenumber} < 15}{
				& \includegraphics[width=28pt, height=28pt]{results/xie/v11/image_00\arabic{imagenumber}.jpg}
			} \\

			DGM 
			\forloop{imagenumber}{1}{\value{imagenumber} < 10}{
				& \includegraphics[width=28pt, height=28pt]{results/xie2/v11/00\arabic{imagenumber}.png}
			}\forloop{imagenumber}{10}{\value{imagenumber} < 14}{
				& \includegraphics[width=28pt, height=28pt]{results/xie2/v11/0\arabic{imagenumber}.png}
			} \\

			DVAE 
			\forloop{imagenumber}{1}{\value{imagenumber} < 10}{
				& \includegraphics[width=28pt, height=28pt]{results/dvae/stgconv/v11/scene000\arabic{imagenumber}.png}
			}\forloop{imagenumber}{10}{\value{imagenumber} < 14}{
				& \includegraphics[width=28pt, height=28pt]{results/dvae/stgconv/v11/scene00\arabic{imagenumber}.png}
			}
		\end{tabular}
	\end{center}
	\caption{Synthesis of dynamic texture \emph{Firepot}}
	\label{fig:firepot}
\end{figure}
	\end{appendices}

%% file: math_commands.tex
\usepackage{amsmath,amsfonts,bm}

\def\figref#1{Fig.~\ref{#1}}

\def\secref#1{Section~\ref{#1}}
\def\subref#1{Subsection~\ref{#1}}

\def\eqref#1{Eq.~(\ref{#1})}

\def\1{\bm{1}}

\def\rvh{{\mathbf{h}}}

\def\rvv{{\mathbf{v}}}

\def\rvx{{\mathbf{x}}}
\def\rvy{{\mathbf{y}}}

\def\vh{{\bm{h}}}

\def\vv{{\bm{v}}}

\def\vx{{\bm{x}}}
\def\vy{{\bm{y}}}

\def\mA{{\bm{A}}}
\def\mB{{\bm{B}}}
\def\mC{{\bm{C}}}

\def\mF{{\bm{F}}}

\def\mI{{\bm{I}}}

\DeclareMathAlphabet{\mathsfit}{\encodingdefault}{\sfdefault}{m}{sl}
\SetMathAlphabet{\mathsfit}{bold}{\encodingdefault}{\sfdefault}{bx}{n}

\def\sM{{\mathbb{M}}}

\newcommand{\R}{\mathbb{R}}

